\documentclass{article}

\PassOptionsToPackage{numbers}{natbib} 

\usepackage[preprint]{neurips_2025}

\usepackage{amsmath}
\usepackage{amssymb}
\usepackage{amsthm}
\usepackage{graphicx}
\usepackage{subcaption}

\usepackage[utf8]{inputenc} 
\usepackage[T1]{fontenc}    
\usepackage{hyperref}       
\usepackage{url}            
\usepackage{booktabs}       
\usepackage{amsfonts}       
\usepackage{nicefrac}       
\usepackage{microtype}      
\usepackage{xcolor}         

\title{Rollout-LaSDI: Enhancing the long-term accuracy of Latent Space Dynamics.}

\author{%
  Robert Stephany \\
  Center for Applied Scientific Computing \\
  Lawrence Livermore National Laboratory \\
  Livermore, Ca 94550 \\
  \texttt{stephany1@llnl.gov}
  \And
  Youngsoo Choi \\
    Center for Applied Scientific Computing \\
    Lawrence Livermore National Laboratory \\
    Livermore, Ca 94550 \\
    \texttt{choi15@llnl.gov} \\
}

\begin{document}
\maketitle

\begin{abstract}
    Solving complex partial differential equations is vital in the physical sciences, but often requires computationally expensive numerical methods.
    Reduced-order models (ROMs) address this by exploiting dimensionality reduction to create fast approximations. 
    While modern ROMs can solve parameterized families of PDEs, their predictive power degrades over long time horizons.
    We address this by (1) introducing a flexible, high-order, yet inexpensive finite-difference scheme and (2) proposing a Rollout loss that trains ROMs to make accurate predictions over arbitrary time horizons.
    We demonstrate our approach on the 2D Burgers equation.
\end{abstract}

\section{Introduction}
\label{sec:Intro}
The profound success of \emph{Deep Neural Networks (DNNs)} \cite{he2016residual, krizhevsky2017alexnet, vaswani2017attention, brown2020gpt3} has given rise to \emph{Scientific Machine Learning (SciML)}, which uses machine learning to address challenges in science and engineering. 
SciML has engendered novel PDE solvers \cite{raissi2019pinn}, algorithms to discover governing models \cite{brunton2016sindy, rudy2017pdefind, rackauckas2020ude}, and \emph{Reduced Order Models} (ROMs) to accelerate simulations \cite{he2023glasdi, bonneville2024lasdi_review}.

Numerically solving \emph{Partial Differential Equations (PDEs)} --- \emph{full-order models (FOMs)} --- is central to the physical sciences \cite{thijssen2007comp_physics} and engineering \cite{jasak2007openfoam, anderson2021mfem}.
However, high-fidelity solvers are computationally expensive \cite{andrej2024mfem_exascale}, hindering their usefulness in time-sensitive applications such as model predictive control \cite{kaiser2018sindy_mpc}. 
ROMs address this by providing fast approximations to FOM solutions \cite{fries2022lasdi}.

A ROM typically comprises: (1) an encoder mapping a FOM state (solution at a specific time) to a low-dimensional \emph{latent space}, (2) \emph{latent dynamics} describing the time evolution of the encoded states, and (3) a decoder returning latent states to FOM states \cite{berkooz1993pod, schmid2010dmd, fries2022lasdi, champion2019sindy_autoencoder}. 
ROMs are predicated on the \emph{manifold hypothesis}: many high-dimensional datasets lie on or near a low-dimensional manifold \cite{gorban2018manifold_hypothesis}. 
If the encoder-decoder pair can approximate a chart and its inverse, they can absorb much of the dataset's complexity, leaving simple latent dynamics to learn.
We can use such a ROM to quickly approximate the FOM solution via the following workflow: (1) encode the initial FOM state, (2) evolve the latent dynamics, then (3) decode to approximate the final FOM state.
Time-stepping occurs in latent space and can be orders of magnitude faster than solving the FOM \cite{bonneville2024gplasdi}.

Recently, several DNN-based ROMs have been proposed.
Champion et al. \cite{champion2019sindy_autoencoder} coupled an autoencoder with SINDy \cite{brunton2016sindy} to encode high-dimensional time series to latent ones that evolve according to simple \emph{ordinary differential equations (ODEs)}. 
The \emph{Latent Space Dynamics Identification (LaSDI)} framework extended this approach to parameterized\footnote{Parameters may alter the initial condition and/or the governing equation.} collections of PDEs \cite{fries2022lasdi}.

LaSDI operates on a collection of time series, each one consisting of FOM states for a specific (and known) parameter value.
LaSDI trains an autoencoder and learns distinct latent dynamics for each parameter instance\footnote{Because parameters change the FOM dynamics, a single set of latent dynamics cannot serve all values.}. 
At inference, LaSDI encodes the initial FOM state and interpolates among the learned latent dynamics to obtain dynamics for a new parameter. 
LaSDI has several extensions \cite{park2024tlasdi, tran2024wlasdi, anderson2025mlasdi, chung2025latent, he2025physics}, including for interpolating the latent dynamics: an intrusive variant leverages known governing equations to greedily add parameters during training \cite{he2023glasdi}; a non-intrusive variant, \emph{GPLaSDI}, uses \emph{Gaussian processes (GPs)} \cite{rasmussen2003gaussian} to add parameters and interpolate dynamics without knowing the FOM \cite{bonneville2024gplasdi, bonneville2023gplasdi_neuralips}. 
This paper builds on \emph{GPLaSDI}.

While GPLaSDI represents significant progress, it faces two limitations that we address in this paper.
First, for time series with nonuniform time steps, GPLaSDI estimates latent derivatives with a two-point finite difference scheme, yielding poor estimates and hindering learning. 
We introduce an inexpensive finite-difference scheme with higher-order accuracy that works on nonuniform time series. 
Second, GPLaSDI, like SINDy, approximates the latent dynamics \emph{locally} by learning latent dynamics that approximately hold at each time step.
Small errors can accumulate, degrading the predictive power of the ROM over long time horizons.
We add a \emph{Rollout loss} that trains the ROM to predict future FOM states over arbitrary horizons. 
These enhancements improve long-horizon prediction accuracy while maintaining the ROM's computational efficiency at inference.

\section{Methodology}
\label{sec:Method}
In this paper, the FOM is a parameterized family of PDEs:
\begin{equation}
\begin{aligned}
    \tfrac{d}{dt} u_{\theta}\left(t, X\right) &= F\left( u_{\theta}\left(t, X\right), t, X, \theta \right), \qquad (t, X) \in (0, T] \times \Omega, \\
    u_{\theta}(0, X) &= u_{0}(X, \theta).
\end{aligned}
\label{eq:PDE}
\end{equation}
Here, $\Omega \subseteq \mathbb{R}^{N_s}$ is the spatial portion of the problem domain, $T > 0$ is the final time, and $\theta \in \Theta \subseteq \mathbb{R}^p$ is a parameter that defines PDE's initial condition and governing equation. 
We assume \eqref{eq:PDE} can be solved by a high-fidelity solver for all $\theta \in \Theta$.
We let $\{\vec{u}_{\theta}(t_j^{\theta})\}_{j=0}^{N_t(\theta)} \subseteq \mathbb{R}^{N_u}$ denote a numerical solution to the FOM \eqref{eq:PDE} at times $0 = t_0^{\theta}\le\cdots\le t_{N_t(\theta)}^{\theta} \le T$ using $N_u$ spatial nodes.
Our training set starts with time series for $N_{\theta} \in \mathbb{N}$ parameter values, $\{\theta_i\}_{i=1}^{N_{\theta}}$.
We assume we can dynamically generate new time series, but do not know \eqref{eq:PDE} (our approach is \emph{non-intrusive}).

Our ROM consists of an autoencoder (with encoder $\varphi_e : \mathbb{R}^{N_u} \to \mathbb{R}^L$ and decoder $\varphi_d : \mathbb{R}^L \to \mathbb{R}^{N_u}$, both of which are trainable DNNs) with latent space $\mathbb{R}^L$ ($L\ll N_u$).
The encoder maps each state, $\vec{u}_{\theta}(t_{i}^{\theta}, X)$, to a low-dimensional latent representation, $\varphi_e(\vec{u}_{\theta}(t_j^{\theta}))$. 
Restricted to the data manifold $M$, we train the autoencoder to satisfy $\varphi_d \circ \varphi_e|_{M} \approx \mathrm{Id}_{M}$ by minimizing the mean absolute error, 
\begin{equation}
    \mathcal{L}_{\text{Recon}} = \frac{1}{N_t} \sum_{i = 1}^{N_{\theta}} \sum_{j = 1}^{N_t(\theta_i)} \left\| \vec{u}_{\theta_i}\left(t_{j}^{\theta_i}\right) - \left( \varphi_d \circ \varphi_e \right)\left( \vec{u}_{\theta_i}\left(t_{j}^{\theta_i}\right) \right) \right\|_1, \quad N_t = \sum_{i = 1}^{N_{\theta}} N_t(\theta_i).
\label{eq:loss:recon}
\end{equation}
The encoder can iteratively map a FOM time-series to the latent space, yielding a latent time series.
Let $\vec{z}_{\theta}(t_j^{\theta}) = \varphi_e(\vec{u}_{\theta}(t_j^{\theta}))$. 
We posit there are unknown, parameter-specific \emph{latent coefficients} - $A_{\theta} \in \mathbb{R}^{L \times L}$ and  $b_{\theta} \in \mathbb{R}^{L}$, such that
\begin{equation}
    \dot{\vec{z}}_{\theta}\left(t_{i}^{\theta}\right) \approx A_{\theta} \vec{z}_{\theta}\left(t_{i}^{\theta}\right) + b_{\theta}, \qquad t \in (0, T].
\label{eq:ld}
\end{equation}
We fit these coefficients by minimizing the \emph{Latent Dynamics loss}, 
\begin{equation}
    \mathcal{L}_{\text{LD}} = \frac{1}{N_t} \sum_{i = 1}^{N_{\theta}} \sum_{j = 1}^{N_t(\theta_i)} \left\| \dot{z}_{\theta}\left(t_{j}^{\theta_i}\right) - \left[ A_{\theta_i} \varphi_{e}\left( \vec{u}_{\theta_i}\left( t_{j}^{\theta_i} \right) \right) + b_{\theta_i} \right] \right\|_2^2. 
\label{eq:loss:LD}
\end{equation}
We zero-initialize $A_{\theta_i}$ and $b_{\theta_i}$ and learn them jointly with $\varphi_e, \varphi_d$ \footnote{Unlike GPLaSDI, which computes latent coefficients via differentiable linear regression at each step, we treat them as trainable parameters for stability.}. 
We learn distinct coefficients for each training parameter value.

Unlike GPLaSDI, we use an $\mathcal{O}(h^2)$ three-point finite difference scheme that supports nonuniform time steps (common in real data and stiff, adaptively stepped systems). 
Let $U$ be a normed vector space, $f:\mathbb{R}\to U$ with $U$, and suppose we know $f(t-a)$, $f(t)$, $f(t+b)$ for some $a, b>0$. 
Then,
\begin{equation}
    f'(t) = \tfrac{-b}{a(a + b)} f(t - a) + \tfrac{b - a}{ab} f(t) + \tfrac{a}{b(a + b)} f(t + b) + \mathcal{O}(h^2), \quad h = \max\{a, b\}. 
\label{eq:fd:central}
\end{equation}
This follows by taking third-order Taylor expansions of $f(t - a)$ and $f(t + b)$ and solving for $c_{-1}, c_0, c_1 \in \mathbb{R}$ such that $c_{-1}f(t - a) + c_0 f(t) + c_1 f(t + b)$ approximates $f'(t)$ with $\mathcal{O}(h^2)$ error \cite{burden1997numerical}. 
We use analogous left/right three-point formulas, giving us one-sided, nonuniform stencils with $\mathcal{O}(h^2)$ accuracy.

Because the coefficients in \eqref{eq:fd:central} are rational functions of $a$ and $b$, we can compute the derivative of an entire time series in linear time; in practice, runtime is just $2\times$ that of the standard three-point scheme. 
The resulting high-fidelity latent-velocity estimates improve our ability to learn and Rollout the latent dynamics, especially when using higher-order ODE solvers (e.g., RK4).

\begin{figure}[hbt!]
    \centering
    \includegraphics[width=\linewidth]{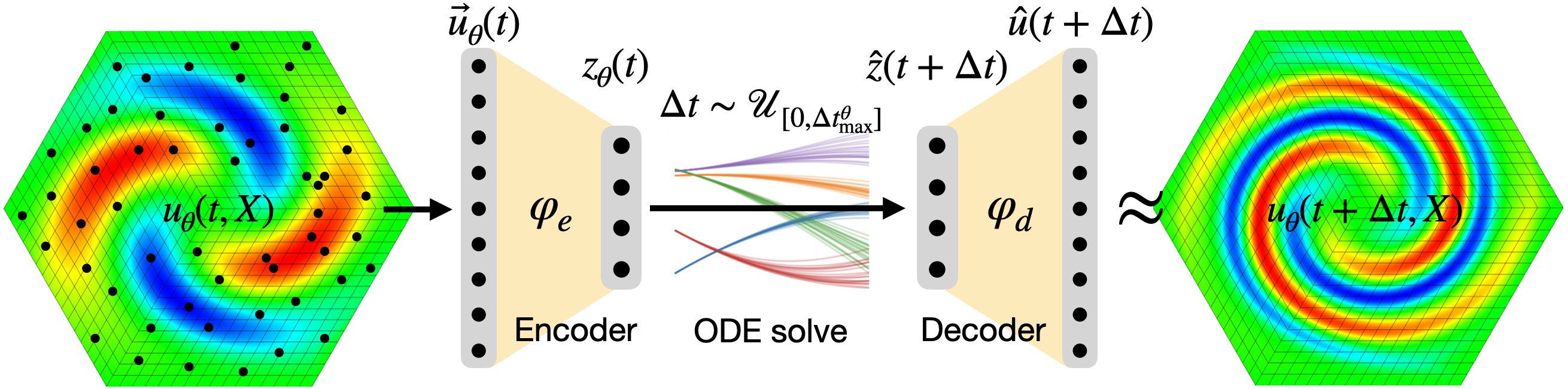}
    \caption{Rolling out $\vec{u}_{\theta}(t)$, the discretization of $u_{\theta}(t, X)$. 
    From left to right, we begin with $\vec{u}_{\theta}(t)$, a numerical approximation of the FOM solution, $u_{\theta}(t, X)$ (black circles represent the spatial nodes). 
    We encode $\vec{u}_{\theta}(t)$, use the encoding as the latent dynamics' initial condition.
    We solve the dynamics over a random time horizon, then decode the final latent state to predict the future FOM state.}
    \label{fig:rollout}
\end{figure}

To improve long-horizon accuracy, we introduce a \emph{Rollout loss}. 
At each epoch and for each parameter $\theta$, we choose a maximum Rollout horizon $\Delta t_{\max}^{\theta}$ (annealed from near zero during training) and define all frames $\vec{u}_{\theta}(t^{\theta})$ with $t^{\theta} + \Delta t_{\max}^{\theta} \le T$ as \emph{rollable}.
We encode each rollable frame, $\vec{u}_{\theta}(t)$, to get a latent vector, $\vec{z}_{\theta}(t)$, which we use as the initial condition to the latent dynamics. 
We then we integrate using a differentiable RK4 solver to a random\footnote{Different frames will be simulated for different amounts of time. This changes each epoch.} future time $t + \Delta t^{\theta}$.
Here, $\Delta t^{\theta} \sim \mathcal{U}\big(0,\Delta t_{\max}^{\theta}\big)$.
We decode the final latent state, $\hat{z}_{\theta}(t + \Delta t^{\theta})$, to get $\hat{u}_{\theta}(t + \Delta t^{\theta}) = \varphi_d\!\big(\hat{z}_{\theta}(t + \Delta t^{\theta})\big)$, and compare it against $\tilde{u}_{\theta}(t + \Delta t^{\theta})$, a cubic-spline interpolation of the FOM solution.
Doing this for each frame and parameter gives us the Rollout loss:
\begin{equation}
    \mathcal{L}_{\text{Rollout}} = \frac{1}{N_{ro}} \sum_{i = 1}^{N_{\theta}} \sum_{j = 1}^{N_{ro}(\theta_i)} \left\|  \tilde{u}_{\theta_i} \left(t_{j}^{\theta_i} + \Delta t_{\text{ro}}^{\theta_i}(j) \right) - \hat{u}_{\theta_i}\left(t_j^{\theta_i} + \Delta t_{ro}^{\theta_i}(j) \right) \right\|_1,
\label{eq:loss:rollout}
\end{equation}
where $N_{ro}(\theta_i)$ is the number of rollable frames for $\theta_i$ and $N_{ro} = \sum_{i = 1}^{N_{\theta}} N_{ro}(\theta_i)$. 
$\mathcal{L}_{\text{Rollout}}$ depends on the encoder/decoder parameters and the latent coefficients. 
Figure~\ref{fig:rollout} illustrates the procedure.

We train our ROM using Adam \cite{kingma2014adam} to minimize the following loss:
\begin{equation}
    \mathcal{L}\left(\varphi_e, \varphi_d, \left\{ A_{\theta_i}, b_{\theta_i} \right\}_{i = 1}^{N_\theta} \right) = \eta_{1}\mathcal{L}_{\text{Recon}} + \eta_{2}\mathcal{L}_{\text{LD}} + \eta_{3}\mathcal{L}_{\text{Rollout}} + \eta_{4}\sum_{i = 1}^{N_{\theta}} \left( \left\| A_{\theta_i} \right\|_{F}^2 + \left\|b_{\theta_i}\right\|_2^2 \right),
\label{eq:loss}
\end{equation}
where $\eta_1, \ldots, \eta_4$ are hyperparameters. 
We adopt GPLaSDI's GP-based coefficient interpolation to adaptively add parameters to the training set; selecting those whose latent-coefficient posteriors induce the largest predictive variance in FOM space.
At test time, we use the GP posterior mean to infer latent coefficients for new parameters \cite{bonneville2024gplasdi}. 
We refer to our algorithm as \emph{Rollout-LaSDI}.
We will make our source code available after review to avoid revealing the authors' identities.

\section{Experiments and Discussion}
\label{sec:Experiment}

We test Rollout-LaSDI on the 2D Burgers equation with periodic \emph{Boundary Conditions (BCs)}:
\begin{equation}
\begin{aligned}
    \dot{u} &= -u u_x - u u_y + \nu \left( u_{xx} + u_{yy} \right) \\
    u\left(0, (x, y) \right) &= \exp\left( -k (x^2 + y^2)\right)\sin\left(\tfrac{\pi}{2} 
    \omega x\right)\sin\left(\tfrac{\pi}{2} \omega y\right).
\end{aligned}
\label{eq:Burgers}
\end{equation}
We consider two parameters: $\nu$, scaling the Laplacian, and $\omega$, controlling the spatial frequency of the initial condition.
We generate FOM solutions using a high-fidelity finite-difference solver on $(0, T] \times \Omega = (0, 2] \times [-2, 2]^2$.
We discretize this domain with $500$ time points and $51 \times 51$ spatial nodes ($N_u = 2601$).
The encoder is a fully connected DNN with widths $[2601,\,250,\,100,\,100,\,100,\,5]$ ($L = 5$) and $\sin$ activation functions; the decoder mirrors the encoder.
Finally, during training, we set $\left\{ \eta_1, \eta_2, \eta_3, \eta_4 \right\} = \left\{ 1, 1, 1, 0.001 \right\}$ (see \eqref{eq:loss}).

\begin{figure}[htbp]
    \centering
    \begin{subfigure}{0.25\textwidth}
        \includegraphics[width=\linewidth]{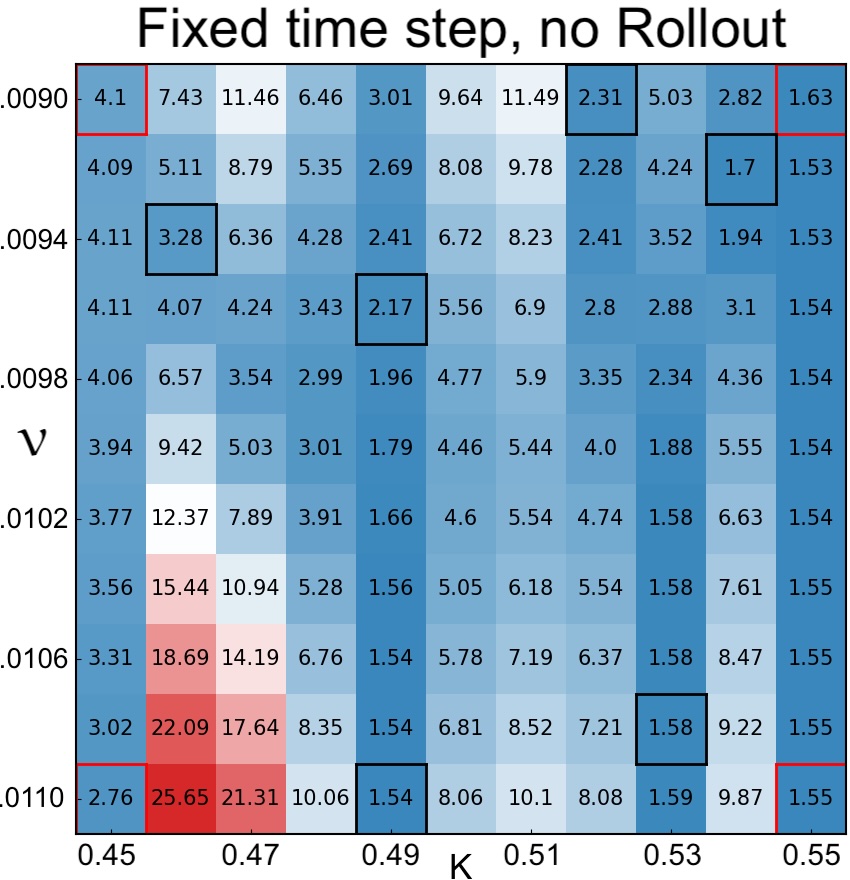}
    \end{subfigure}
    \hfill
    \begin{subfigure}{0.23\textwidth}
        \includegraphics[width=\linewidth]{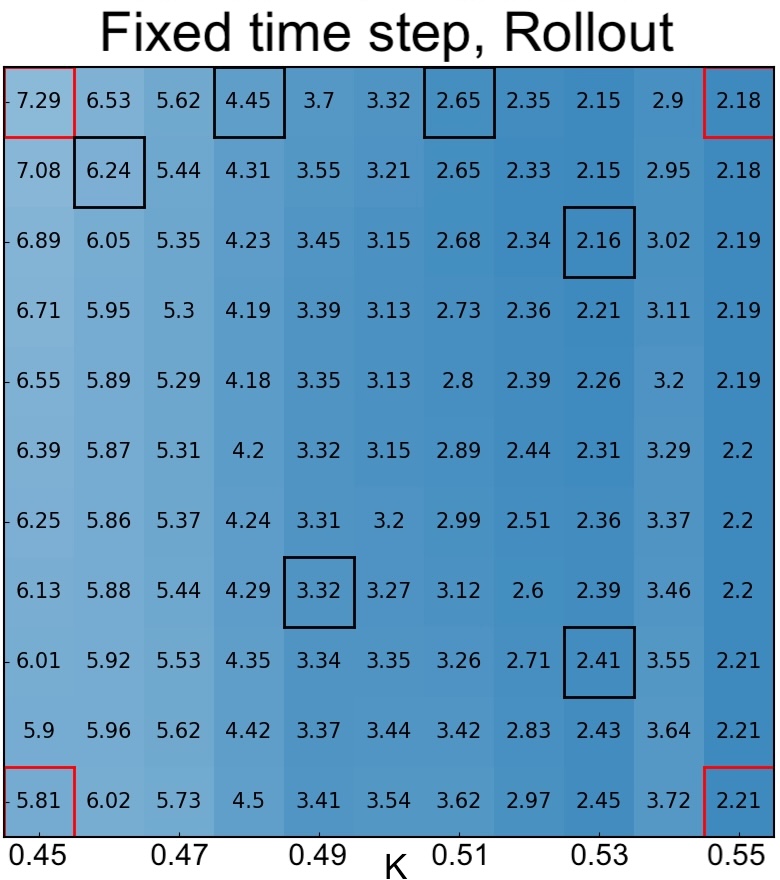}
    \end{subfigure}
    \hfill
    \begin{subfigure}{0.23\textwidth}
        \includegraphics[width=\linewidth]{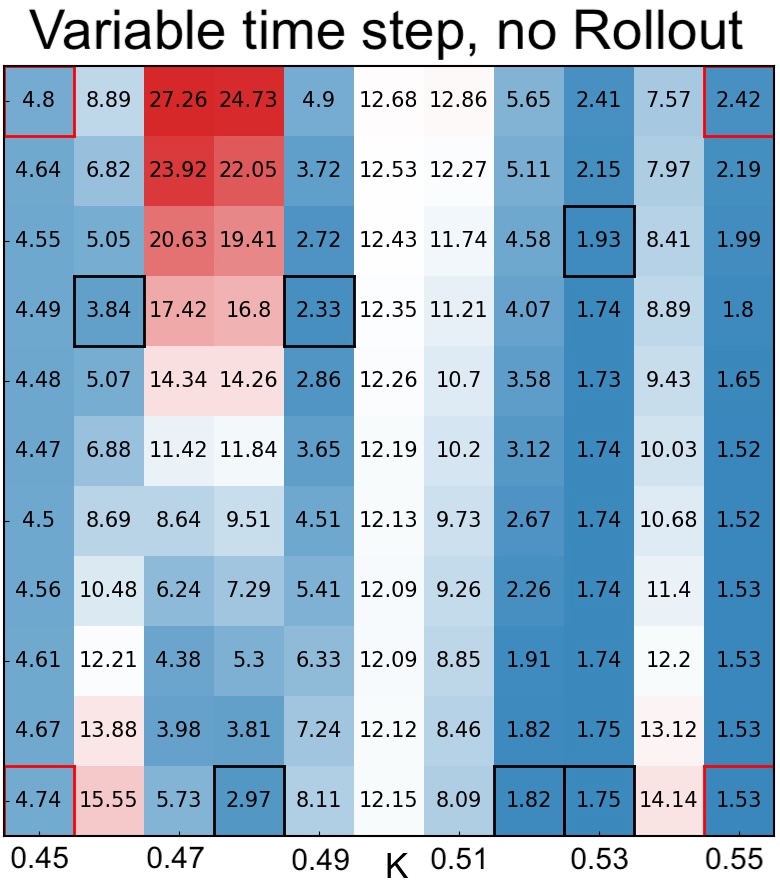}
    \end{subfigure}
    \hfill
    \begin{subfigure}{0.255\textwidth}
        \includegraphics[width=\linewidth]{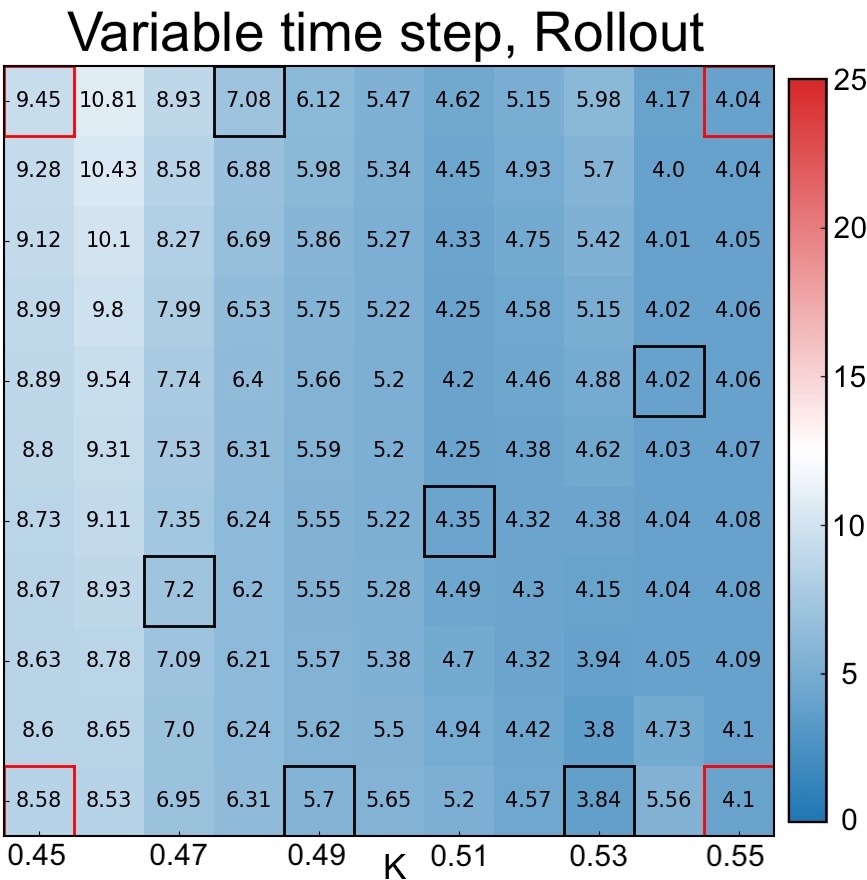}
    \end{subfigure}
    
    \caption{Error for all $121$ parameter values with and without Rollout. Here, the four red cornered boxes represent the parameter values that are originally in the training set, while the black ones represent parameters that were dynamically added to the training set during training.}
    \label{fig:Heatmaps}
\end{figure}

We train for $17,500$ epochs using Adam (lr $=10^{-3}$), perform greedy sampling (to pick a $\theta$ and corresponding time-series to add to the training set) every $2500$ epochs, and use $20$ GP samples to estimate FOM variance \cite{bonneville2024gplasdi}.
The test set consists of an $11 \times 11$ grid ($121$ parameter combinations, with $11$ values for both $\nu$ and $\omega$).
The training set begins with $4$ parameters (red-bordered squares in Fig.~\ref{fig:Heatmaps}) but dynamically grows (black-bordered squares) via greedy sampling as the ROM trains \cite{bonneville2024gplasdi}.

At test time, the final ROM and learned GPs predict the FOM solution for each $\theta$: we encode the FOM initial condition\footnote{During testing, this is the only FOM information used by the ROM.}, set latent coefficients to the GP posterior mean, integrate the latent dynamics on $(0,2]$, and decode to obtain the predicted FOM time series.
Because Rollout-LaSDI only uses $\mathcal{L}_{\text{Rollout}}$ during training, and because Runge–Kutta solvers (used in GPLaSDI and Rollout-LaSDI) handle variable step sizes, inference in Rollout-LaSDI is identical to Rollout with GPLaSDI.

To measure the performance of the final ROM, we compute the maximum relative error between a FOM frame, $\vec{u}_{\theta}(t_j)$, and Rollout-LaSDI's prediction, $\hat{u}_{\theta}(t_j)$:
\begin{equation}
    \text{Error}(\theta) = \text{max}_{k} \left( \frac{\text{mean}_{i} \left| \vec{u}_{\theta}(t_k)_i - \hat{u}_{\theta}(t_k)_i \right|}{\sigma_{i, j} \left\{ \left|\vec{u}_{\theta}(t_i)_j \right| \right\}} \right).
\label{eq:Error}
\end{equation}
We normalize by the standard deviation of FOM frame components to avoid division by zero, which could occur with node-specific or time-specific normalizations near the boundary of $\Omega$ or when the FOM state approach zero.

We run four experiments: two with variable time steps (each step size randomly varied) and two with fixed steps. 
In each subset, we train one ROM with Rollout loss and one without it (the two ROMs and their training procedure are identical otherwise). 
Figure~\ref{fig:Heatmaps} reports the results across all $\theta$ values.

In both fixed- and variable-time-step experiments, adding Rollout reduces the maximum relative error by $3\times$ and the median error by $2\times$.
Experiments with variable time steps have slightly higher relative error than those with a fixed time step size, as expected.
These results demonstrate that Rollout improves the long-horizon accuracy of the final ROM without harming test-time performance.

Beyond accuracy improvements, our enhancements maintain GPLaSDI's computational efficiency. 
Our ROMs can achieve the same $10^5$-factor speedup as GPLaSDI \cite{bonneville2023gplasdi_neuralips, bonneville2024gplasdi} while providing superior long-horizon predictions.
We find Rollout and nonuniform time-stepping crucial for robust, real-world ROMs; this study shows that both enhancements strengthen state-of-the-art methods like GPLaSDI.

\begin{ack}
RS was supported by the Sydney Fernbach Postdoctoral Fellowship under LDRD number 25-ERD-049. 
YC was supported by the U.S. Department of Energy (DOE), Office of Science, Office of Advanced Scientific Computing Research (ASCR), through the CHaRMNET Mathematical Multifaceted Integrated Capability Center (MMICC) under Award Number DE-SC0023164.
Livermore National Laboratory is operated by Lawrence Livermore National Security, LLC, for the U.S. Department of Energy, National Nuclear Security Administration under Contract DE-AC52-07NA27344. LLNL document release number: LLNL-CONF-2010452.
\end{ack}

\bibliographystyle{plainnat}
\bibliography{Bibliography}

\end{document}